\documentclass{article}

\usepackage{arxiv}

\usepackage[utf8]{inputenc} 
\usepackage[T1]{fontenc}    
\usepackage{url}            
\usepackage{booktabs}       
\usepackage{nicefrac}       
\usepackage{enumitem}   
\setlist[itemize]{leftmargin=*}
\setlist[enumerate]{leftmargin=*}
\usepackage{natbib}         
\setcitestyle{authoryear,open={[},close={]}}
\usepackage{graphicx}        
\usepackage{subcaption}
\makeatletter
\renewcommand*\subcaption@label{%
  \caption@withoptargs\subcaption@@label}
\makeatother

\graphicspath{ {images/} }
\usepackage{color}
\usepackage{multicol}
\usepackage[hypcap=false]{caption}

\makeatletter
\newenvironment{Figure}
{\def\@captype{figure}}
{}
\makeatother

\usepackage{hyperref}       

\usepackage{appendix}
\title{FairyTailor: A Multimodal Generative Framework for Storytelling}

\author{
  Eden Bensaid \\
  IBM Research, MIT CSAIL\\
  \texttt{edenbd@mit.edu} \\
   \And
   Mauro Martino \\
  IBM Research\\
   \texttt{mmartino@us.ibm.com} \\
      \And
         Ben Hoover \\
  IBM Research\\
   \texttt{benjamin.hoover@ibm.com} \\
   \And
 Hendrik Strobelt \\
  IBM Research, MIT CSAIL\\
  \texttt{hendrik.strobelt@ibm.com} \\
}

\begin{document}

\maketitle
\begin{abstract}

Storytelling is an open-ended task that entails creative thinking and requires a constant flow of ideas. Natural language generation (NLG) for storytelling is especially challenging because it requires the generated text to follow an overall theme while remaining creative and diverse to engage the reader. 

In this work, we introduce a system and a web-based demo, FairyTailor\footnote{available at \url{https://fairytailor.org}}, for human-in-the-loop visual story co-creation. Users can create a cohesive children's fairytale by weaving generated texts and retrieved images with their input. FairyTailor adds another modality and modifies the text generation process to produce a \textbf{coherent} and \textbf{creative} sequence of text and images. To our knowledge, this is the first dynamic tool for multimodal story generation that allows interactive co-formation of both texts and images. It allows users to give feedback on co-created stories and share their results. We release the demo source code\footnote{\url{https://github.com/EdenBD/MultiModalStory-demo}} for other researchers' use.   
\end{abstract}

\begin{multicols}{2}
\section{Introduction}

Automated story generation strives to generate compelling stories automatically \citep{text-storytelling}. A story consists of  sentences describing a series of events. Story generation introduces compelling challenges to existing Natural Language Generation models. Compared to more constrained text generation tasks, such as machine translation and summarization, which follow existing content, story text generation has an open-ended nature. It requires diversity and creativity while adhering to a continuous narrative.

Multimodal content is prevalent in social media posts, news articles, and commercials. Among the audio, videos, and pictures modalities, images are the most common modality to accompany textual content. Adding images can enrich the content and catch readers' attention. Therefore, semi-automatically generating a multimodal story can produce more attractive results, especially for young readers, and augment short stories. 

An interactive writing platform can support writers by suggesting new ideas and continuing previous content. It can offer exciting and entertaining directions that are nevertheless relevant to the writer's work. Moreover, it can alleviate writers' inertia and keep them motivated and involved in writing. Giving writers full editing power to control the final story's content keeps the users engaged. 

\begin{figure*}[ht]
\centering
\includegraphics[width=0.9\linewidth]{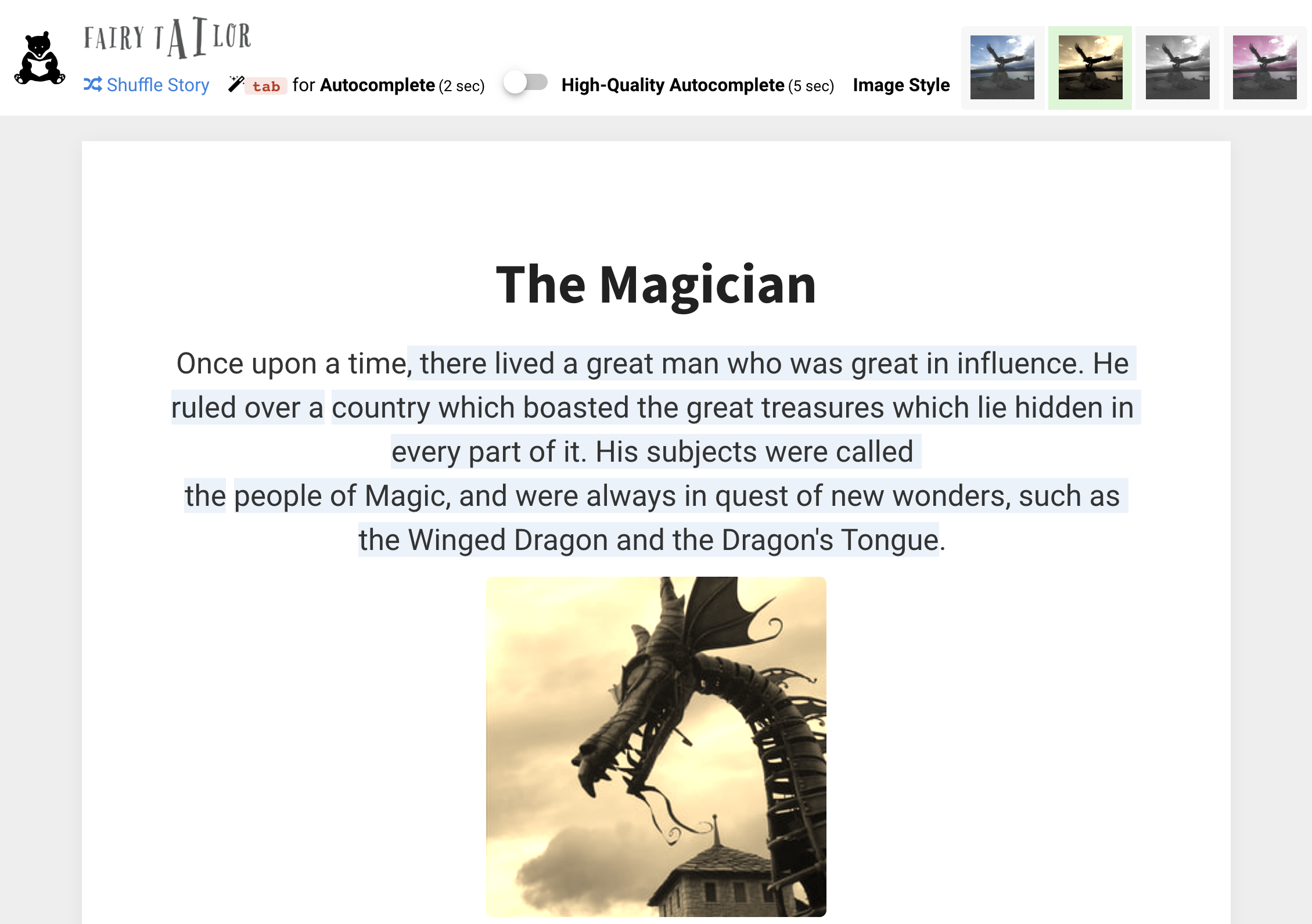}
\captionof{figure}{FairyTailor user's co-created story. User's text has no background, versus machine-generated text has a blue background.}
\label{fig:demo-example}
\end{figure*}


A challenging story generation aspect is sustaining long-term memory and producing coherent text within an overall theme \citep{nlg-challenge}. Another major challenge is that while adhering to a general theme and tone, stories must evolve their composition and progress to new directions. 

Current storytelling models are limited to focus on text modality \citep{text-storytelling, planandwrite, infilling, re-ranker, storium}, without incorporating another modality such as images. Training generative models on a single type of input, task and domain often results in a lack of generalization and robustness \citep{gptt2, compositional-learning}. Multimodal generative models, which are capable of relating and sharing information across multiple modalities \citep{multimodal}, can create a representation that focuses on objects and the relations among them \citep{compositional-learning}. Existing vision-and-language models \citep{vilbert, compositional-learning} are not trained for storytelling generation but for other downstream tasks, such as image captioning \citep{vilbert}, alignment prediction \citep{vilbert}, masked multimodal learning \citep{vilbert}, bounding box prediction \citep{compositional-learning} and visual relation prediction \citep{compositional-learning}. 

To encourage the model to produce more abstract representations, overseeing the textual content generation with another modality can yield promising results. There is a compelling need for a multimodal system that incorporates both to create an engaging story.


In this work, we propose a method for multimodal story co-creation based on a sequential language-and-vision framework. We apply our method and demonstrate the co-formation of visual tales through an interactive, web-based public demo. Finally, we provide early evidence through a qualitative user study of the proposed platform on the story-generation objective.

\section{Related Work}

There are several previous work methods for controllable story generation \citep{text-storytelling, planandwrite,infilling, re-ranker} that aim to produce coherent text with an appealing plot. Topic conditioned models produce stories from a compact topic input \citep{text-storytelling, planandwrite}. The advantage of the topic encoding is that it can create a concise, progressing storyline \citep{planandwrite}. However, since the Seq2Seq model tends to focus on recently generated text and specific parts of the prompt, the plot derails from the storyline within the few (3-5) generated sentences \citep{planandwrite}. It also frequently generates similar sentences without any sense of progression \citep{text-storytelling}. Storyline conditioned models propose tighter conditioning during the story generation by continuously directing the start to a specific ending \citep{infilling, re-ranker}. The \textit{Unsupervised Hierarchical Story Infilling} \citep{infilling} conditions the language model on keywords that are probable to appear between the beginning and the ending of the story. The \textit{Narrative Interpolation for Generating and Understanding Stories} \citep{re-ranker} generates several candidates and re-ranks them to take the one with the best overall coherence. Our approach mixes ideas from both topic and storyline controlled models by augmenting extracts from the dataset with automatically generated keywords and continuously re-ranking the text generation.

Story visualization architectures retrieve \citep{neural-retrival} or generate \citep{storygan} images to illustrate a given story, i.e. a multi-sentence paragraph. \textit{Coherent Neural Story Illusration} (CNSI) \citep{neural-retrival} suggests an encoder-decoder framework that can retrieve a coherent sequence of images from visualGenome \citep{visualGenome} by predicting images' feature representations from encoded sentences and parse tree extractions \citep{neural-retrival}. StoryGAN can generate a coherent sequence of images dependent on the text by concatenating the current sentence with contextual information vector encoded from the entire story \citep{storygan}. Other text-to-image generation models such as BigGAN \citep{bigGAN}, stackGAN \citep{stackGAN} and Dall-E \citep{dalle} evaluate their results on various text sources. As detailed in subsection \ref{why-retrieval}, we favor image retrieval since they align better with user interaction times and our free-form textual content was not descriptive enough to generate valuable images. We retrieve images independently according to story pieces and use a broad image dataset that corresponds to our intended stories' genre. 

Even though there are no multimodal architectures for storytelling that we are aware of, combining vision and language for a joint representation is addressed by several successful models. MVAE \citep{mvae} consists of one VAE model that assumes conditional independence of modalities to use product-of-experts (PoE) and reduce the number of parameters. The VAEGAN model \citep{compositional-learning} uses a VAE for text and a GAN for images on a modified multimodal objective that minimizes variational divergences \citep{compositional-learning}. As part of the pretrain-then-transfer approach, ViLBERT \citep{vilbert} aims to serve as a common platform for visual grounding. It has two separate streams for visual and textual inputs that interact through co-attentional transformer layers \citep{vilbert}. 

Previous approaches to generate stories suffer from repetition \citep{text-storytelling}, are unable to consistently condition on a theme \citep{planandwrite} and struggle to produce a grounded, evolving storyboard \citep{rl-story-generation, re-ranker, infilling}. Story visualization often requires specific, informative text \citep{storygan}, and takes significantly longer than image retrieval on our one-GPU compute resource to create relevant images. To address these problems in story text generation and story visualization, we offer a multimodal story generation platform that collaborates with writers. A similar interactive writing platform is STORIUM \citep{storium}, an online collaborative storytelling community. However, it is intended for text completions of long stories that follow the STORIUM narrative format \citep{storium}. 

Our proposed multimodal story-generating framework aims to generate \textbf{creative} and \textbf{coherent} short tales by taking advantage of \textbf{multimodal} robust representation of stories, \textbf{decoder-based transformer} architecture \citep{gptt2, schema-guided} and \textbf{controllable text generation}. Multimodal frameworks have been proven successful over their unimodal counterparts on various downstream tasks  \citep{compositional-learning, vilbert}. Transformer models such as GPT-2 \citep{gptt2}, GPT-3 \citep{gpt3} and TransformerXL \citep{transformerxl} have successfully used decoder transformer blocks \citep{transformer-decoder} to generate diverse, stable text. Controllable generation have encouraged generation of coherent texts \citep{text-storytelling, planandwrite,infilling, re-ranker}. Therefore, it seems promising to compare and evaluate our multimodal generative framework on the complex story-generation task.

\section{Data Collection and Data Wrangling}\label{subsection:datasets} 
In this section, we describe the datasets and pre-processing used for creating the benchmark system and the final design. We evaluated several sources of data for each modality. A detailed data analysis is available in Appendix \ref{appc}.

\subsection{Text datasets}
To help finetuning models, we tried data from an open-source dataset and from a manually collected text collection. Reddit WritingPrompts \citep{text-storytelling} is an open-source dataset that provides a writing prompt before each story. The stories are varied in their subjects, language and writers. The manually collected dataset consists of public domain children's books from Project Gutenberg that are suitable for young readers. These books are hand-picked and cleaned before use.

To fit Reddit WritingPrompts data to our text generation style and adjust it to the transformer model pre-requisites \citep{gptt2}, we pre-process and clean the data to: trim stories to 1000 words, clean special characters and symbols, remove offensive words, filter stories that were classified as having a negative sentiment, merge prompt and corresponding story to one pair and add end-of-sentence between them and at the end of the story.

From approximately 300K stories, we train the benchmark model on 35K prompt-and-story pairs that were predicted to express positive sentiment. To predict the tonality of a given story, we use a pre-trained BERT \citep{bert} with an added GRU layer that is fine-tuned on the IMBD dataset for a sentiment analysis regression task. The model returned a value from 0-1, representing extremely negative to positive sentiment. The selected stories have a sentiment score above 0.9.

We handpick Gutenberg books relevant to fairy tale generation (complete list available at Appendix \ref{appa}). We further clean the data by splitting stories into 500 tokens extracts, removing redundant new lines, metadata, offensive language, and special characters, and adding a generated prompt before each extract to keep a prompt-story structure and encourage controllable coherent generation.

We use approximately 9K 500-token extracts to fine-tune the benchmark model for the second time, after Reddit WritingPrompts, and to fine-tune the final model for the first and only time. We do not train the final model on Reddit WritingPrompt because of the unpredictable nature of stories. Even after filtering stories, many were unsuitable for our intended young audience.  

\subsection{Image datasets} 

For image retrieval, we assessed several open-source resources to find a varied dataset that includes sceneries, people, and animals that are more closely related to fairy tales. After evaluating COCO \citep{coco-captioning}, Unsplash \citep{unsplash} and Flickr30k \citep{flickr-dataset} caption-image datasets, we use Unsplash due to images relative objects' diversity and landscapes nature.

\section{System Architecture}
We present two evolving architectures that we used in the process of this work. Our benchmark model \ref{subsection:banchmark} introduced multimodal generation but suffered from repetition, inconsistency, and negative sentiments. We mitigated those flaws in our final model \ref{subsection:final-model} by improving the framework and changing the data. 

\subsection{Benchmark Design}\label{subsection:banchmark}

The benchmark model shown in Figure \ref{fig:benchmark}generates text and accordingly retrieves images. The architecture introduces a novel multimodal element. The images guide the text generation process by re-ranking the generated story samples by how coherent and relevant the retrieved images are. Another feature is the automatic text ranking performed after text generation that tests the generations' readability, diversity, and sentiment.  


We perform two fine-tuning rounds of the GPT-2 model \citep{gptt2} using the huggingface library \citep{hugginface}. The first is on Reddit WritingPrompt \citep{text-storytelling} to fine-tune the model to a prompt-story template. The second is on our individually collected children's books dataset to adapt the model to a younger audience. 

 To encourage creativity while maintaining consistent text generation, we tested the \textit{top-k random sampling} method ($k = 50$) that was used in the \textit{Hierarchical Neural Story Generation} model (with $k = 10$) \citep{text-storytelling}, but the results were repetitive. We decided to use Nucleus Sampling \citep{decoding} instead.

For image retrieval, the benchmark architecture extracts frequent nouns from the generated text to retrieve corresponding images from Flickr30K \citep{flickr} caption-image dataset. 

\begin{Figure}
      \centering
\includegraphics[width=\linewidth]{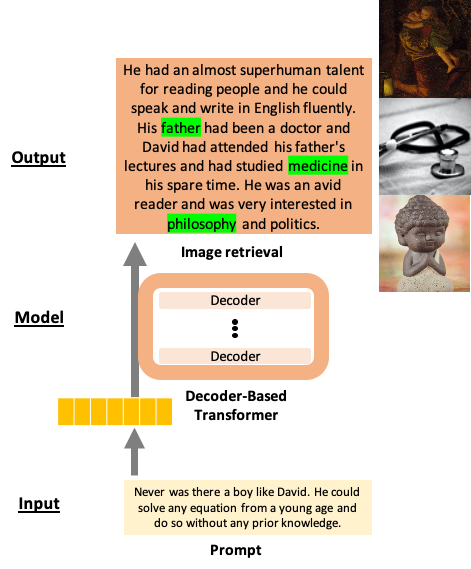}
\captionof{figure}{Benchmark Model Architecture. The system generates text from a given prompt by using a fine-tuned decoder-based transformer, and then retrieves images from Flickr dataset \citep{flickr} according to key nouns.}
\label{fig:benchmark}
\end{Figure} 

\subsection{Final Design}\label{subsection:final-model} 

Testing the benchmark model revealed a few flaws that we approach in our final model. First, The text completions are often repetitive, incoherent, inappropriate, and dark. Second, the independently retrieved images are inconsistent (e.g., might get a different female figure each time). To improve upon the benchmark generation, we revise both the text generation and the image retrieval methods. The final model is shown in Figure \ref{fig:final}.

\subsubsection{Text Modality} 
We added several re-ranker metrics to significantly increase the ranker's role and score texts according to their readability, positiveness, diversity, simplicity, coherency, and tale-like manner. The re-ranker computes the min-max normalization (\ref{eq:min-max}) to rescale each feature across all generated texts so that all features contribute equally.
        \begin{equation}\label{eq:min-max}
        scaled\_scores = \frac{scores - \min(scores)}{\max(scores) - \min(scores)}  
        \end{equation}
Furthermore, we increased the re-ranker frequency. To maintain a coherent text generation, we re-rank after each end-of-sentence token. By re-ranking, we only keep the better half of the generation and filter out the rest. 
The following text measures contribute to the re-ranking:

\textbf{Readability} calculates the length of sentences and length of words to estimate how complex the text is.
    \begin{equation}\label{eq:readibility}
    readability = 0.5*word\_chars + 
    sent\_words
    \end{equation}
Where \textit{word\_chars} and \textit{sent\_words} are equal to -10 if number of words and number of sentences are zero respectively. The 0.5 multiplier gives a higher rank to the number of words per sentence.

\textbf{Positive Sentiment} uses SentiWordnet \citep{SentiWordnet} to compute the positivity polarity. SentiWordnet assigns sentiment scores to each WordNet \citep{wordnet} synonym group. WordNet is popular for information retrieval tasks and does not require pre-training. Since we do not have a supervised sentiment dataset for tales, SentiWordNet predictions were more accurate than neural nets trained on different datasets.

\textbf{Diversity} calculates the fraction of unique words from the total number of words.
\begin{equation}\label{eq:diversity}
diversity = \frac{\textnormal{len}(\textnormal{set}(filtered\_words))}{\textnormal{len}(filtered\_words)}
\end{equation}
\textit{filtered\_words} are word tokens that exclude stop words (e.g., at, in, is) and punctuation. The score is equal to zero if there are no filtered words.

\textbf{Simplicity} calculates the fraction of tale-like characteristic words in the given text.
\begin{equation}\label{eq:simplicity}
simplicity = \textnormal{len}(\textnormal{set}(filtered\_words) \cap freq\_words)
\end{equation}
\textit{freq\_words} are precalculated to represent seven precent of the most frequent words in the collected Gutenberg fairy tales corpus. 

\textbf{Coherency} calculates the Latent Semantic Analysis (LSA) similarity within the story sentences compared to the first sentence. The calculation includes three steps:
\begin{enumerate}
    \item Computing the LSA embedding of the tf-idf document-term matrix per sentence. \newline $embeddings = embedder(text\_sentences)$
    \item Computing the pairwise cosine similarity for each sentence against all other sentences.\newline
    $similarity = cosine\_similarity(embeddings)$
    \item Computing the final similarity score by comparing the first sentence to the rest of the sentences: 
    sum(\textit{similarity}[0][1:]).
\end{enumerate}

\textbf{"Tale like"} computes the KL divergence loss between a preset GPT-2 and a fine-tuned GPT-2 generated texts' prediction scores. A higher score is better since it usually implies that the text is more similar to the fine-tuned distribution and different from the preset GPT-2 distribution. The computation consists of the following steps:
\begin{enumerate}
    \item Tokenizing and encoding the text to \textit{tokens\_ids} to prepare it for the forward pass. 
    \item Computing the logits of the present model \textit{logists\_preset} and of the fine-tuned model \textit{logits\_finetuned} with forward pass on \textit{tokens\_ids}.
    \item Returning the difference score according to the KL-diveregence loss of the two models logits: \newline
    torch.nn.KLDivLoss(logSoftmax(\textit{logists\_preset}), \newline softmax(\textit{logits\_finetuned})).
\end{enumerate}
    
\subsubsection{Image Modality} \label{why-retrieval}
We evaluated three open-source implementations for text to image synthesis,  BigGAN\footnote{\url{https://github.com/huggingface/pytorch-pretrained-BigGAN}} \citep{bigGAN}, stackGAN\footnote{\url{https://github.com/hanzhanggit/StackGAN-Pytorch}} \citep{stackGAN} and Dall-E\footnote{\url{https://github.com/lucidrains/DALLE-pytorch}} \citep{dalle}. To generate images with bigGAN, we extracted image categories (i.e., animal and object names) from the generated text. However, bigGAN's limitation to 1000 categories \citep{bigGAN} prohibited the generation of relevant scenes. StackGAN and Dall-E accept any text input. However, since the text is usually unstructured and not descriptive of a scene, the resulting images were often noisy and distorted (Figure \ref{fig:noisy-GAN}). Furthermore, image generation times were significantly longer than image retrieval. One GPU image generation time ranged from 4-30 seconds per image versus image retrieval, which took 0.5-2 seconds per image. 

\begin{Figure}
          \centering
    \includegraphics[width=0.9\linewidth, height=0.3\linewidth ]{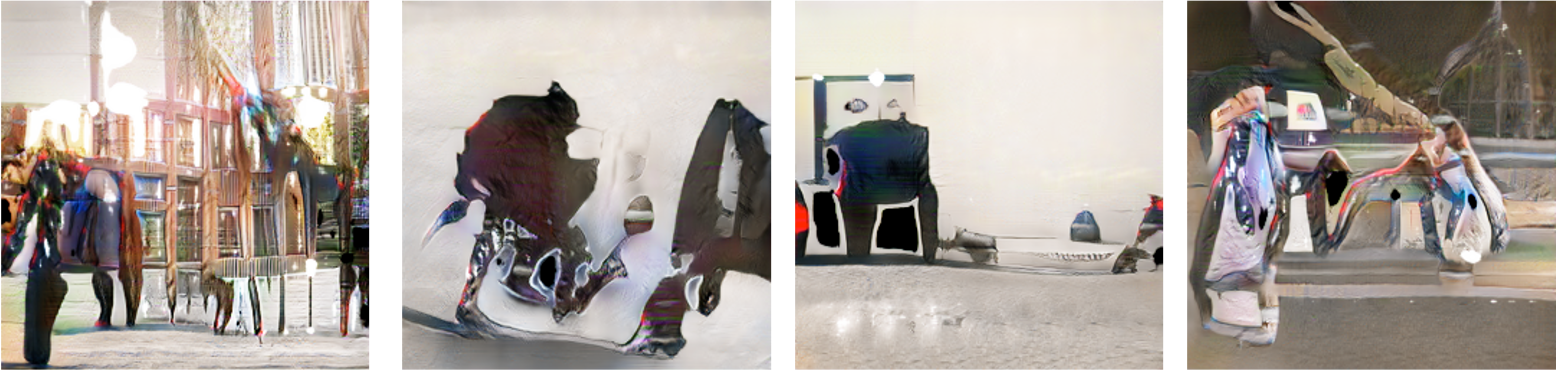}
    \captionof{figure}{StackGAN's generations for the caption: “several men standing outside of small airplane with man retrieving luggage from cart."}
    \label{fig:noisy-GAN}
    \end{Figure} 
    
Therefore, we remained with the image retrieval method, but with a different dataset and an improved retrieval method. We use \citep{clip} to transform the images and generated text to the same latent space. Per-retrieval, we compute the cosine similarity of the text embeddings and the images embeddings. The computation returns the images' ids of the highest-scoring images. The retrieval searches over 2M Unsplash pre-computed embeddings to find the best match to the query\footnote{\url{https://github.com/haltakov/natural-language-image-search}}. 

To achieve a coherent look of story images we fine-tune a neural style transfer model \citep{fast-style-transfer} on several target images shown in Figure \ref{fig:styles}. 

\begin{Figure}
      \centering
\includegraphics[height=0.4\linewidth, width=0.8\linewidth]{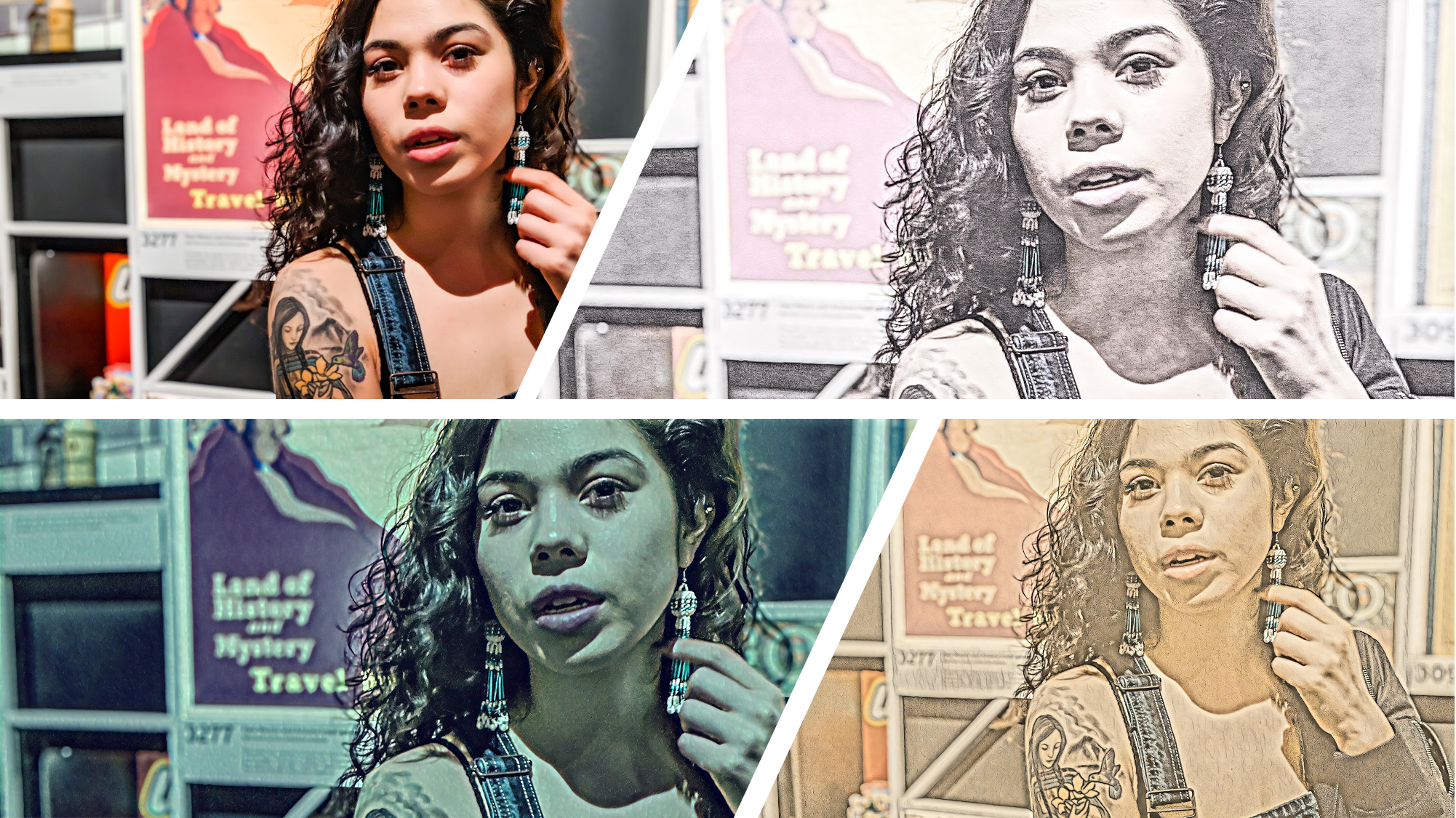}
\captionof{figure}{Style-transfer results on the top-left original image.}
\label{fig:styles}
\end{Figure} 

To improve overall story generation, the framework can generate multiple stories and rank them by their images' consistency. The images' consistency metric is calculated by summing the KL divergence of ResNet \cite{resnet} classification predictions of image pairs. A lower score indicates a smaller difference, which is better. 

\begin{Figure}
      \centering
\includegraphics[width=\linewidth]{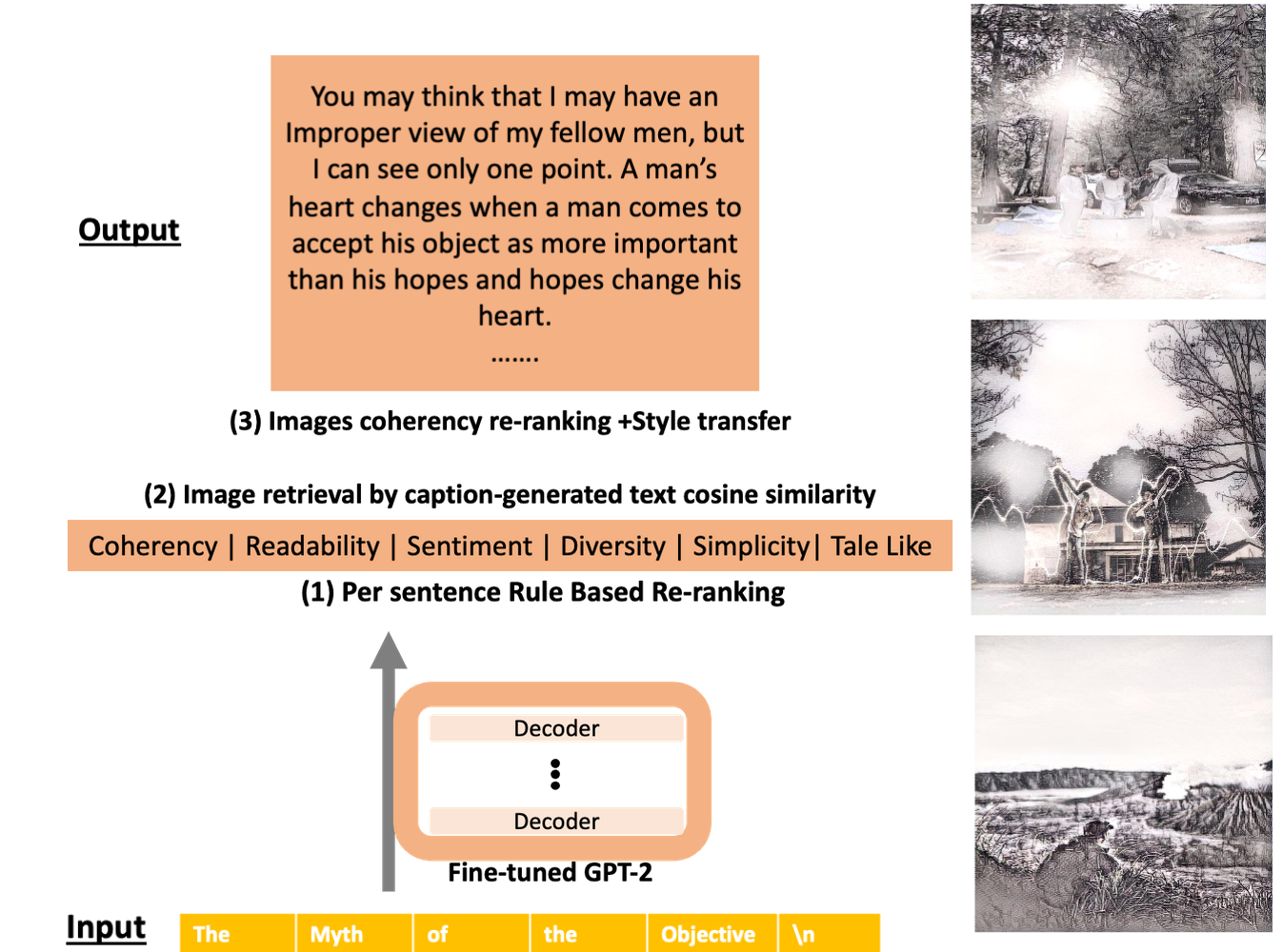}
\captionof{figure}{Final Model Architecture. The system generates text while re-ranking, retrieves images from Unsplash dataset, applies style transfer \citep{fast-style-transfer} and then re-ranks stories according to the story's visual consistency.}
\label{fig:final}
\end{Figure} 

\section{Prototype "FairyTailor"}

FairyTailor is a user interface to access the final multi-modal framework and allow story co-creation. A human writer can start in multiple ways: from scratch, by using a random story primer, or by entering minimal content such as a story title. Fairytailor then offers various modes of autocomplete to assist the writer. 

The web platform skips some of the final design steps to allow for fast user interaction. We use CSS transforms instead of the style transfer neural model, and we do not evaluate stories by the consistency of their image (since we generate one story at a time).

\textbf{Autocomplete.} The faster, more straightforward text autocomplete immediately returns the three completions generated by the fine-tuned model. It may generate empty or irrelevant completions.

\textbf{High-Quality Autocomplete.}
Instead of generating three text completions, the framework generates ten texts, ranks them, and returns the top three. The framework scores texts according to their readability, positiveness, diversity, simplicity, coherency, and tale-like manner. The scoring metrics are detailed in sub-section \ref{subsection:final-model}. 

\begin{Figure}
    \centering
        \begin{subfigure}[Autocomplete]{\linewidth}\includegraphics[width=\linewidth]{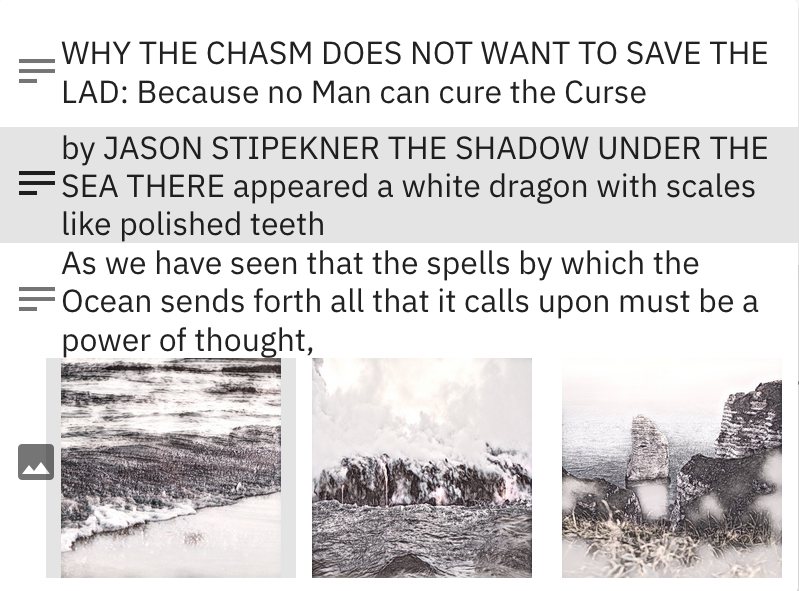}\caption{Autocomplete}\end{subfigure}\quad
    \begin{subfigure}[High-Quality Autocomplete]{\linewidth}\includegraphics[width=\linewidth]{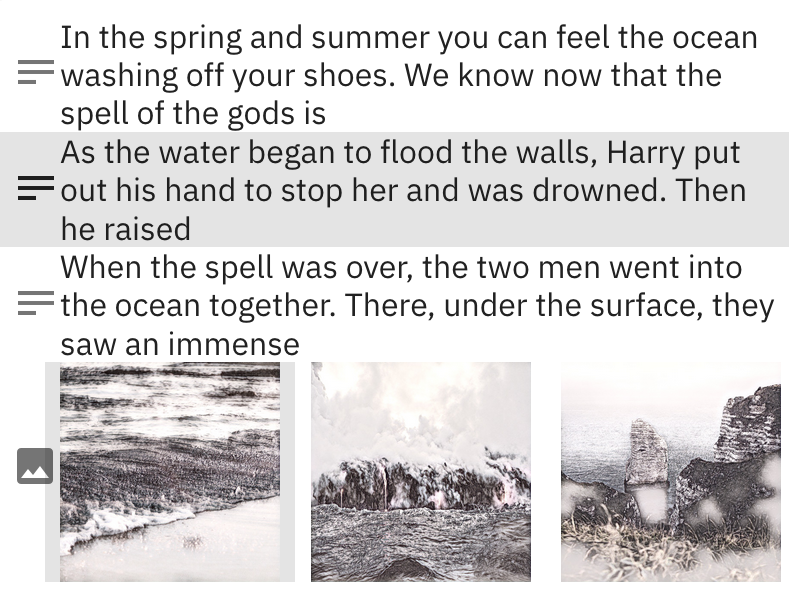}\caption{High-Quality Autocomplete}\end{subfigure}
    \caption{To balance the trade-off between quality and model-inference time, FairyTaylor provides two modes of autocompletion. }
    \label{fig:autocompletions}
\end{Figure}

\textbf{Human vs AI edits.}
Writers can add, delete, and edit the generated text and images as they wish. The generated text is marked differently than user inputted text for data collection and evaluation purposes.

\subsection{FairyTailor as user testing platform}
Creative Natural Language Generation lacks reference texts and heavily relies on user evaluations instead of automatic metrics for quality checks. The demo provides a user-testing platform to share work with others and discover useful patterns quickly. Users can share their experience with a submission form, which will record their ratings, free-form feedback, and story's HTML. Researchers can use the HTML to review aspects of the generated story, such as the ratio of generated vs. user-inputted text and number of images.

\section{Experiment, Evaluation \& Discussion}

Since we used a custom, newly collected dataset and our demo's goal is story co-creation, the most suitable evaluation practice is human evaluations \citep{controllable-reward, storium}. Automated metrics such as Perplexity \citep{perplexity}, BLEU \citep{bleu} and BLEURT \citep{bleurt} are unsuitable to measure creativity and coherence without reference texts. 

We use the FairyTailor prototype and a questionnaire to solicit feedback on the user interface and the generated stories. The feedback questions expand on the evaluation form on the website to understand the user's journey until they freeze and submit their story. We further analyze the users' published stories to verify the efficacy of the generations. We check the ratio of generated vs. user inputted text and the text to image ratio.

\subsection{Qualitative Evaluation}\label{section:methods}

We collected feedback on the demo with story generation experts that have done similar work. The demo's added value from their experienced perspective is summarized in Appendix \ref{appf-evaluations}.  

In addition, we performed structured human evaluations with thirteen students and professionals (six males and seven females). The interviewees included two undergraduate students in Biology Engineering and Computer Science, a Computer Science graduate student, five Computer Science Ph.D. students, two Natural Language Processing researchers, a global marketing executive, a software engineer, and a technical product manager.

The questionnaire form can be found in Appendix \ref{appb}. It is split into two parts to evaluate: (1) \textit{storytelling background} which checks whether the user has written stories before, and in what context and (2) \textit{user feedback} on the generated story (e.g., ranking the story's flow and quality), the interface (e.g., the use of autocomplete versus high-quality autocomplete and the use of images), and the overall experience (e.g., asks what the user liked and did not like). 

The generated story questions are based on \textit{Predicting Generated Story Quality with Quantitative Measures} \citep{story-eval-questions}. They were designed for automated story evaluations and were previously tested on similar tasks.

\subsection{Feedback and Results}
Participants' insights revealed what is enjoyable and what should be improved. The participants commented on the strengths and weaknesses of the interface and the multimodal framework completions after playing with the platform for a 5-10 minutes.  

\subsubsubsection{\textbf{Demo Interface.}}
Overall, the participants found the demo highly engaging. A few users mentioned that the short text autocompletes that are not full sentences encouraged them to press autocomplete again, and they were less likely to delete those shorter completions. 

\subsubsubsection{\textbf{Autocomplete Versus High-Quality Autocomplete.}}
The autocomplete and high-quality autocomplete options demonstrate the difference between the ablated version of the framework and the iterated, final one. Users who tried both indicated that the high-quality autocompletes, which take advantage of the final framework, are significantly better and were willing to wait longer for generations. It emphasizes the benefits that the final model's modifications provide.

\subsubsubsection{\textbf{Multimodal Framework Completions.}}
A third of the submitted stories did not include images because users found them irrelevant to the story, or they did not think their style fits the story they had in mind. The same happened with text completions. If the tone, vocabulary, or ideas that the completion suggested did not fit what the users had in mind, they did not incorporate it. 

\subsubsubsection{\textbf{Published Stories.}}
The average scores of published co-created stories are clarity: 3.4, coherence: 3.5, creativity: 3.7 out of 5. Most of the participants highlighted the beneficial creativity of the platform and its fairy tail-y nature. 


\subsection{Discussion}

We found that people are excited about interactive writing and enjoyed prompting autocomplete. Some of the supportive comments include: "I have a big interest in literature, so this is very fun" and "I love the highly engaging, very polished user interface". Flexibility was key for an enjoyable experience. People liked having control over the content, the placement of the texts and images and the timing of the completions.  
\subsubsection{Strengths}
Since our goal is co-creation of stories with human-in-the-loop, many questions were on the demo user interface. Users highlighted the ease of use and the design of the platform. It is best to use the platform when users are open-minded. One of the users mentioned: "Though I did not know where my story was going initially, the autocomplete helped me find a direction". Prompting autocompletes is likely to generate different suggestions each time, thus helping writers guide the story. 

\subsubsection{Weaknesses}
Some users did not use images because their style or content did not fit their stories. The image retrieval is restricted to the images dataset and thus do not fit every scenario. 
Most users indicated they declined to use the suggested (simple) autocompletion 50\%-75\% of the times they prompted it because it was repetitive or did not fit their motif. However, for users who used the high-quality autocomplete, the numbers were significantly lower, ranging from 0-25\% of the times declining suggestions. This shows, that spending time on content quality matters, and that faster high-quality models are key to future success.

\subsubsection{Improvements}

During the interviews we also collected features on how we could help improving the user experience with FairyTailor overall.
\begin{itemize}[leftmargin=*]
    \item Adding other modes of user interaction, such as changing the image style-transfer style. 
    \item Adding endings completions, since regular autocompletes do not try to summarize or direct the current story. 
    \item A leaderboard of the highest-scoring published stories. 
    \item An option for user-provided text examples to fine-tune the language model and adapt the style and probable vocabulary to the users' intended writing style. 
\end{itemize}

\section{Conclusion and Future Work}

We find that participants enjoyed engaging with FairyTailor to co-create a variety of stories and would use such systems again. FairyTailor is especially beneficial for beginning writers, who find it hard to start and do not envision a specific storyboard in mind. Users mentioned that the completions' creativity helped them find a direction and maintain a continuous flux of ideas. 

The platform is helpful for user testing and automated story evaluations. When the user submits a story, the platform saves its content along with an outlined feedback form. Researchers can quickly evaluate the ratio of generated versus inputted text and inserted images ratio with simple analytics. The platform is publicly available\footnote{\url{https://github.com/EdenBD/MultiModalStory-demo}} for other researchers to deploy their work and user-test a story generation model quickly.

The image modality is novel among other story generation platforms. The images add a touch to the story and are especially prevalent in children's books. When image completions are relevant, users tend to incorporate them. Published stories that included images were ranked higher overall; users who used images praised the images' role in improving their co-generated story's quality. 

From conducting our experiments, we think that the canon of future work in multimodal story generation should include the following ideas and challenges:

\textbf{User-specific completions.} Currently, the autocomplete function is the same for all users. It only changes according to content. However, users' writing style and goals vary. When users have a specific storyboard in mind, the platform might never get what they envisioned and generate irrelevant completions.  Incorporating an interactive feedback loop can mitigate this problem. The deletions or unused autocompletes can guide the model to the users' intentions and produce user-centered results. 

    \textbf{Storyboard completions.} Currently, autocompletes do not explicitly follow a storyboard and are not designated for the beginning, middle, or ending of the story. Suppose a user indicates a need to end or evolve the story by providing a goal-driven storyboard in advance or signaling while writing. In that case, it will be beneficial to have directed autucompletions that follow these cues. 
    
    \textbf{Image generation versus image retrieval.} For generality purposes, it is valuable to generate images according to input, assuming that generated images will be of high quality as retrieved images are and rendered in a suitable time frame.
    
    \textbf{Faster is better.} A technical challenge for interactive human-AI multi-modal co-creation is not only to develop models of higher text/image quality but also models of fast inference time to allow interactive rates during the creation process.

\subsection{Acknowledgments}
This work would not have been possible without the continuous support and feedback from MIT Professor Jacob Andreas. We also thank FairyTailor users who dedicated time to be interviewed and shared helpful insights.  
\end{multicols}
\bibliography{references}
\newpage
\appendix
\appendixpage
\section{Collected Gutenberg Stories' Titles}\label{appa}

The Happy Prince, Andersens Fairy Tales, The Blue Fairy Book, The Adventures of Pinocchio,
Myths Retold by Children,Household Tales, Indian Fairy Tales, Fairy Tales Second Series, MERRY STORIES AND FUNNY PICTURES,
                   Childhoods Favorites and Fairy Stories,The Wonderful Wizard of Oz, Celtic Tales, Our Children, The Little Lame Prince, 
                   The Prince and Betty, The Adventures of Sherlock Holmes,Peter Pan,The Secret Garden,
                    The Jungle Book, The Adventures of Tom Sawyer, A Little Princess, Little Women, Just So Stories, Moby Dick, Treasure Island, The Idiot, A Tale of Two Cities, My Man Jeeves, Sense and Sensibility, The Time Machine, Comic History of the United States, The Velveteen Rabbit, The Book of Dragons, The Snow Image, The Magical Mimics in Oz, Folk Tales from the Russian, Snow-White or The House in the Wood, Dramatic Reader for Lower Grades, A Christmas Hamper, Aesop Fables, My Fathers Dragon, The Peace Egg and Other tales, Indian Why Stories, Folk-Tales of the Khasis, The Paradise of Children, Wonder Stories, The Best American Humorous Short Stories, Hindu Tales from the Sanskrit, The Tale of Johnny Town-Mouse, The Little Red Hen, East of the Sun and West of the Moon, Among the Forest People, True Stories of Wonderful Deeds, English Fairy Tales, Simla Village Tales Or Folk Tales from the Himalayas, Japanese Fairy Tales, Plain Tales of the North, The Wind in the Willows, The Louisa Alcott Reader. A Supplementary Reader for the Fourth Year of School, A Wonder Book for Girls and Boys, Tanglewood Tales, The Pig Brother and Other Fables and Stories, The Worlds Greatest Books, Vol 3, Goody Two-Shoes, The Marvelous Exploits of Paul Bunyan, Christmas Every Day and Other Stories, The Childrens Book of Thanksgiving Stories.

\section{FairyTailor User Test Template}\label{appb}
FairyTailor, available at fairytailor.org, is a visual story co-creation platform created by MIT \& IBM. 
Users can create a cohesive story by weaving automatically generated texts and retrieved images with their input.
\newline
*Required
\begin{enumerate}
    \item Email Address*
    \item Have you written stories before? If yes, elaborate on the intended audience and the stories' structure*
    \item Paste the URL of your story (created after pressing "submit story" at the bottom)*
    \item Do you agree with the following statement?* (Choose one of strongly Disagree, Somewhat Disagree,Neither Agree nor Disagree, Somewhat Agree, Strongly Agree).\cite{story-eval-questions}
    \begin{itemize}
        \item Autocompletes exhibit CORRECT GRAMMAR
        \item Autocompletes occur in a PLAUSIBLE ORDER
        \item Autocompletes MAKE SENSE given sentences before and after them.
        \item Autocompletes AVOID REPETITION
        \item Autocompletes use INTERESTING LANGUAGE
        \item This story is of HIGH QUALITY.
        \item This story is ENJOYABLE.
        \item This story follows ONE OVERALL THEME.
    \end{itemize}
    \item When prompted, how often did you decline to use the suggested autocompletions?* 
    \begin{itemize}
        \item Never
        \item For 25\% of completions
        \item For 50\% of completions
        \item For 75\% of completions
        \item Always
    \end{itemize}
    \item Did you use autocomplete (Tab) or High-Quality autocomplete (shift + Tab)?*
      \begin{itemize}
        \item Mostly autocomplete (Tab)
        \item Mostly HQ autocomplete (shift + Tab)
        \item Both
        \item Other: \_\_
    \end{itemize}  
    \item Please elaborate on your choice above*
    \item Did you use images? Why?*
    \item What did you like?*
    \item What not so much?*
    \item Other comments/ suggestions?
\end{enumerate}

\section{Data Analysis}\label{appc}
The datasets were analyzed to validate their diversity. Each text dataset was inspected to look at the number of sentences (Figure \ref{fig:number-of-sentences}), the Part-Of-Speech (POS) tagging (Figure \ref{fig:pos}) and the least frequent (Figure \ref{fig:least-freq-words}) and most frequent (Figure \ref{fig:most-freq-words}) words. 

The number of sentences' distributions in Figure \ref{fig:number-of-sentences} verifies that our dataset mostly includes shorter stories as the ones we aim to produce. The children's stories corpus has a higher quantity of longer stories, leading to a higher mean of 71 sentences per children's story versus Reddit's mean of 48 sentences.  

The Part-Of-Speech (POS) tagging distributions in Figure \ref{fig:pos} displays higher concentrations of verbs and nouns than adjectives, as expected. Interestingly, in both stories' corpora, the mean number of adjectives is approximately 35\% of the mean number of verbs or nouns. That high ratio might reflect the vivid, artistic nature of storytelling. 

The least frequent words' distributions in Figure \ref{fig:least-freq-words} shows that the Reddit corpus has a more diverse vocabulary than the children's stories corpus. Almost 50\% of Reddit's vocabulary consists of infrequent words in comparison to 40\% in children's stories. 

The most frequent words displayed in Figure \ref{fig:most-freq-words} word clouds show the characteristic old-fashioned style of children's stories. Old, upon, and shall are among the most frequent words in fairy tales, whereas terms like world, people, eyes are common in Reddit stories. Prevalent words such as I, time, and you appear in both corpora. 

\begin{Figure}
    \centering
        \begin{subfigure}[B]{.5\textwidth}\includegraphics[width=\linewidth]{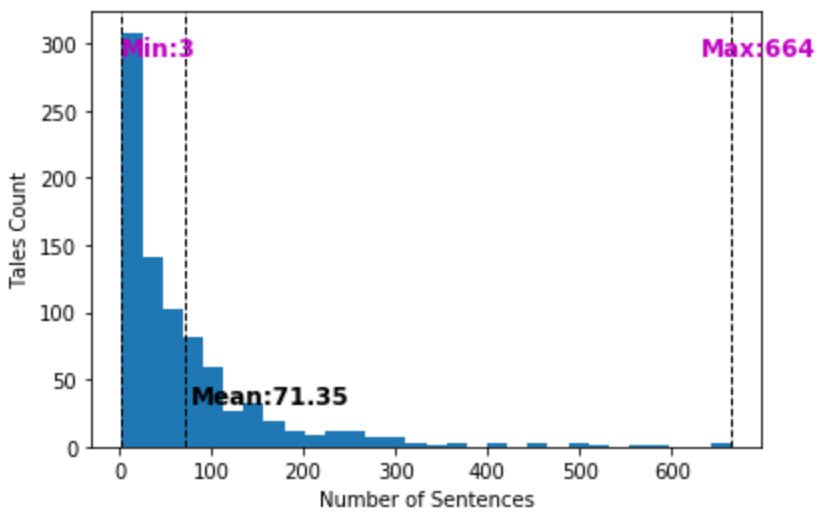}\caption{Children's Stories}\end{subfigure}\hfill
    \begin{subfigure}[B]{.5\textwidth}\includegraphics[width=\linewidth]{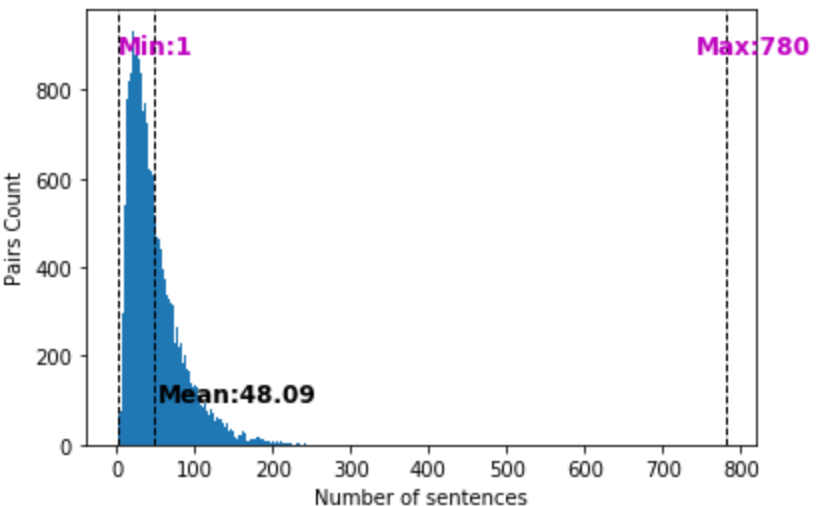}\caption{Reddit Prompt-Story Pair}\end{subfigure}%
    \caption{Number of sentences in text datasets.}
    \label{fig:number-of-sentences}%
\end{Figure}

\begin{Figure}
    \centering
        \begin{subfigure}[B]{.5\textwidth}\includegraphics[width=\linewidth]{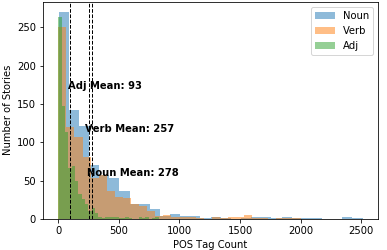}\caption{Children's Stories}\end{subfigure}\hfill
    \begin{subfigure}[B]{.5\textwidth}\includegraphics[width=\linewidth]{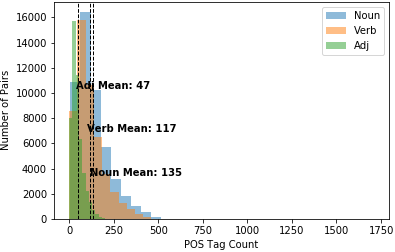}\caption{Reddit Prompt-Story Pair}\end{subfigure}%
    \caption{Part of speech (POS) tagging of Verb, Noun and Adjective (Adj) in text datasets.}
    \label{fig:pos}
\end{Figure}

\begin{Figure}
    \centering
        \begin{subfigure}[B]{.5\textwidth}\includegraphics[width=\linewidth]{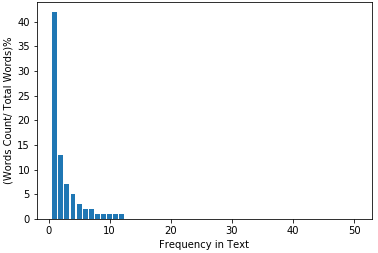}\caption{Children's Stories}\end{subfigure}\hfill
    \begin{subfigure}[B]{.5\textwidth}\includegraphics[width=\linewidth]{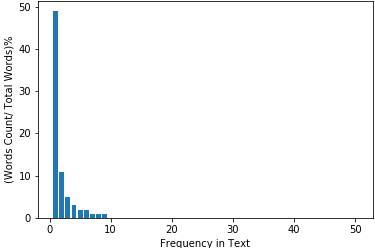}\caption{Reddit Prompt-Story Pair}\end{subfigure}%
    \caption{Frequency of the least frequent words in text datasets.}
    \label{fig:least-freq-words}%
\end{Figure}

\begin{Figure}
    \centering
        \begin{subfigure}[B]{.5\textwidth}\includegraphics[width=\linewidth]{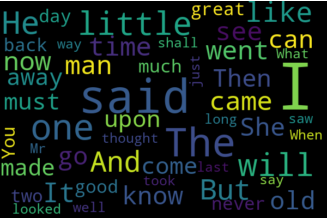}\caption{Children's Stories}\end{subfigure}\hfill
    \begin{subfigure}[B]{.5\textwidth}\includegraphics[width=\linewidth]{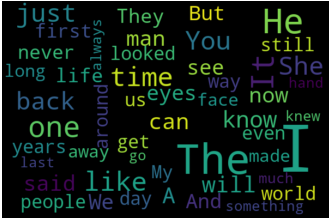}\caption{Reddit Prompt-Story Pair}\end{subfigure}%
    \caption{50 Most frequents words in text datasets.}
    \label{fig:most-freq-words}%
\end{Figure}

\newpage
\section{Benchmark and Final Model Generated Stories}
\begin{Figure}
      \centering
\includegraphics[height=0.6\linewidth, width=\linewidth]{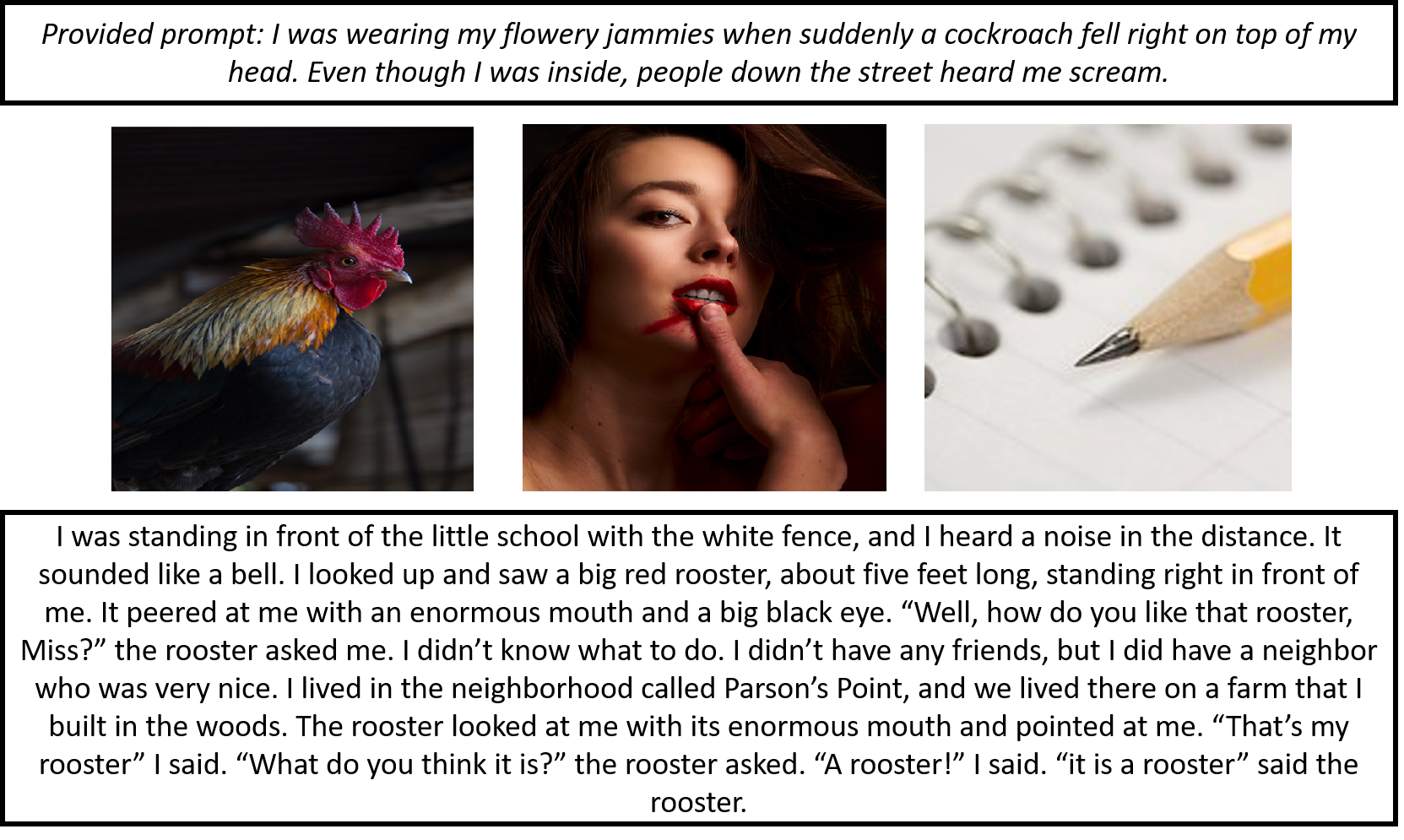}
\captionof{figure}{Benchmark model: Example of a generated story}
\label{fig:benchmark-ex}
\end{Figure} 

\begin{Figure}
      \centering
\includegraphics[height=0.6\linewidth, width=1.1\linewidth]{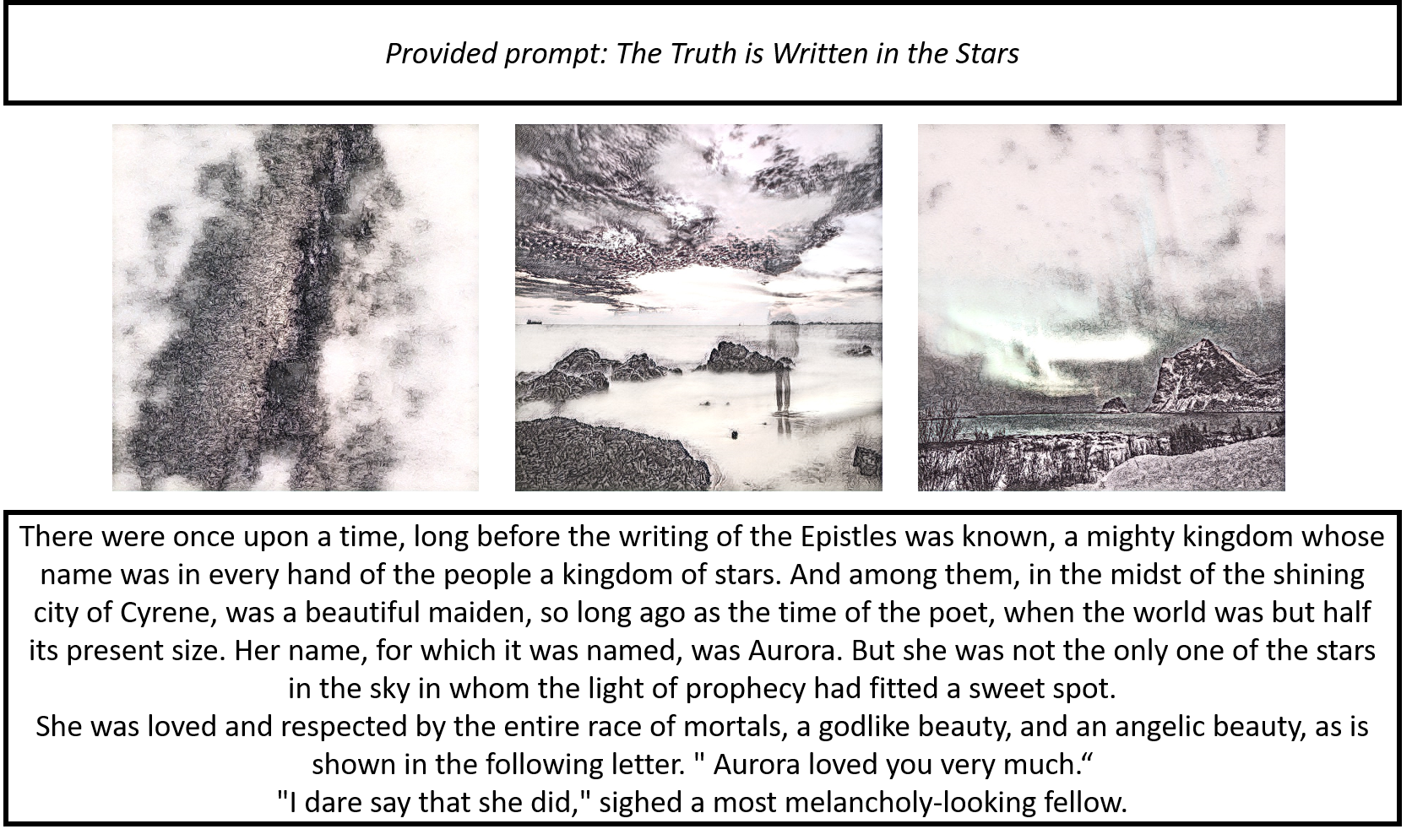}
\captionof{figure}{Final model: Example of a generated story}
\label{fig:final-ex}
\end{Figure}

\newpage
\section{FairyTailor Screenshots}\label{appe-screenshots}

\begin{Figure}
    \centering
        \begin{subfigure}[B]{\textwidth}\includegraphics[width=\linewidth]{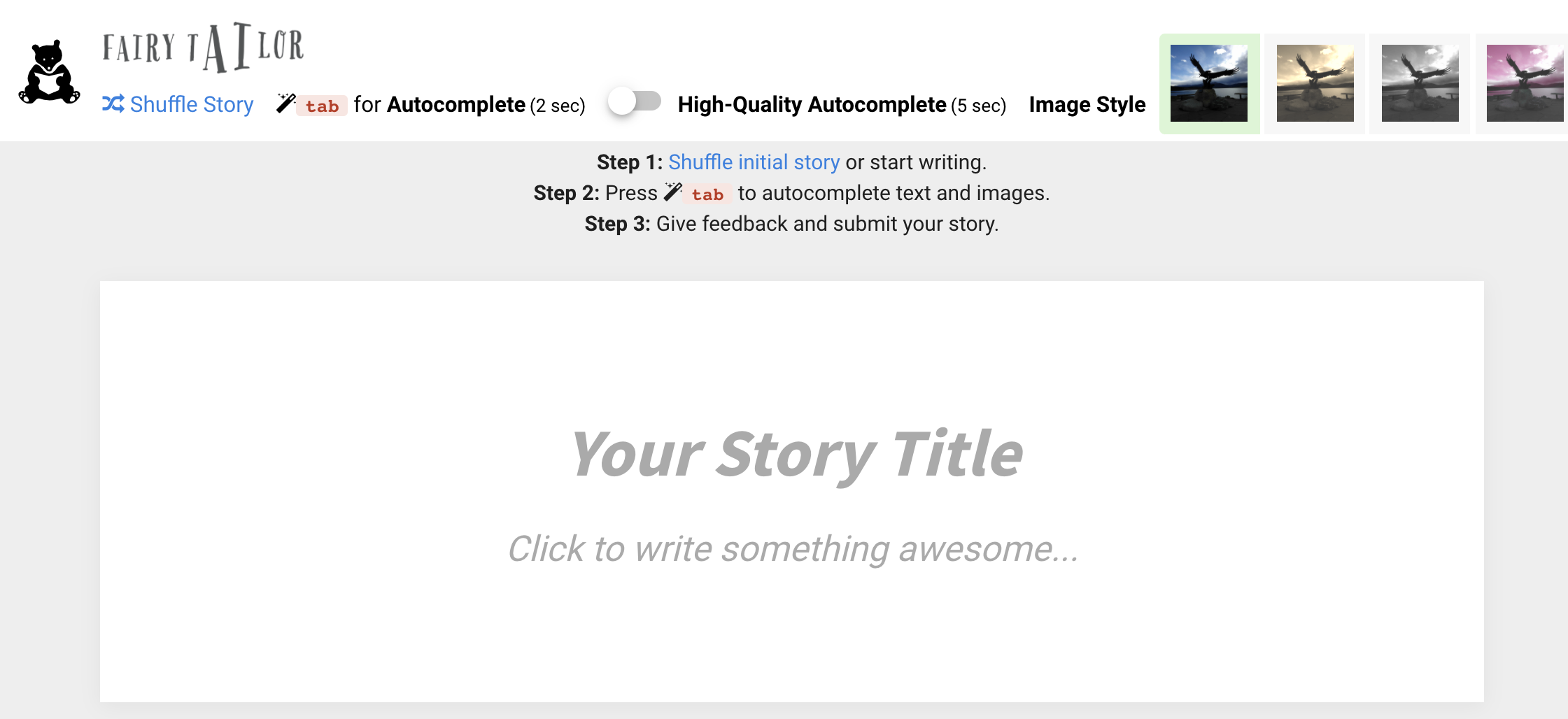}\caption{Landing Page}\end{subfigure}
    \begin{subfigure}[B]{\textwidth}\includegraphics[width=\linewidth]{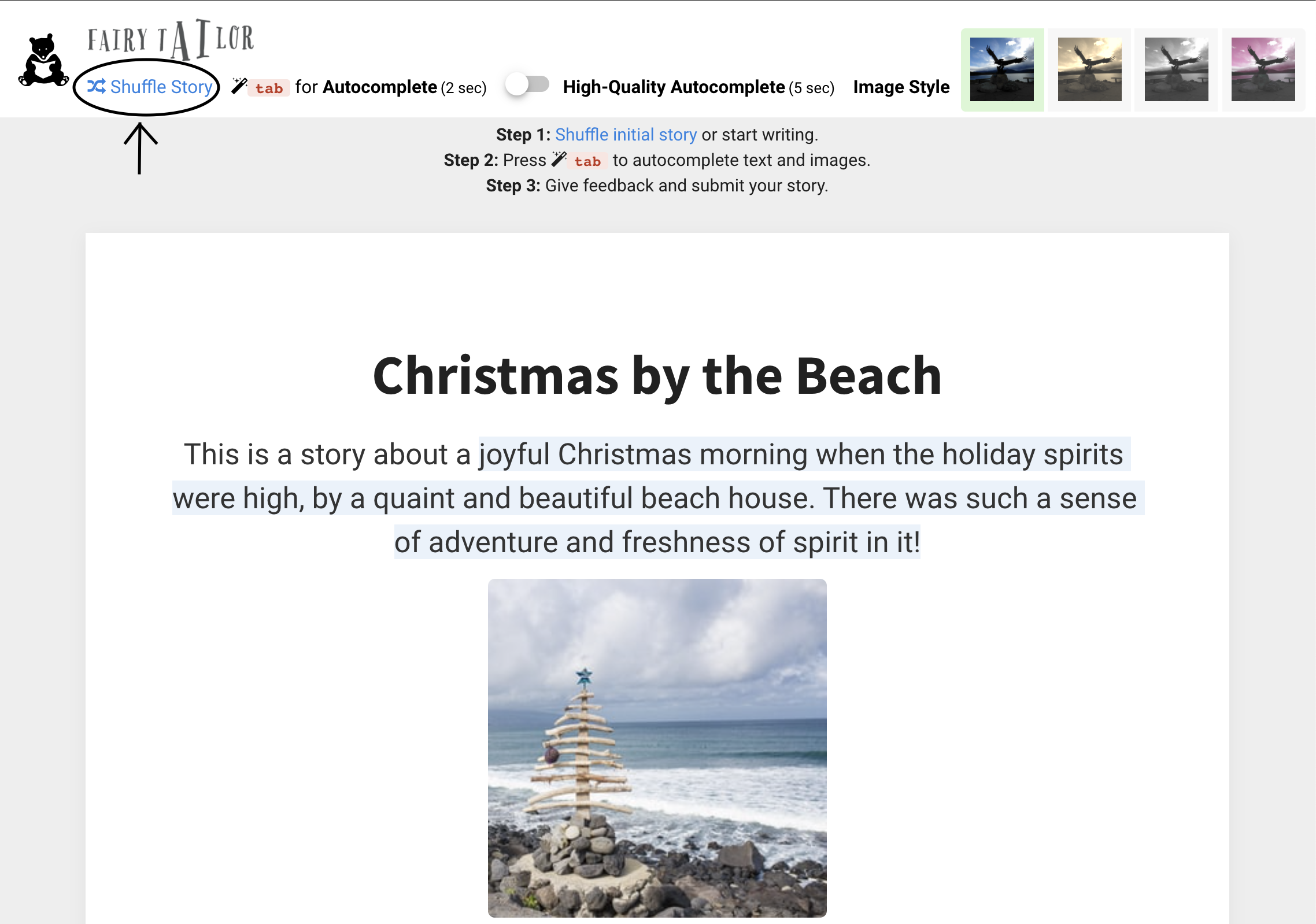}\caption{Random Preset Story}\end{subfigure}%
    \caption{Users can start writing from scratch or use preset examples.}
    \label{fig:pages}%
\end{Figure}

\begin{Figure}
    \centering
        \begin{subfigure}[B]{\textwidth}\includegraphics[width=\linewidth]{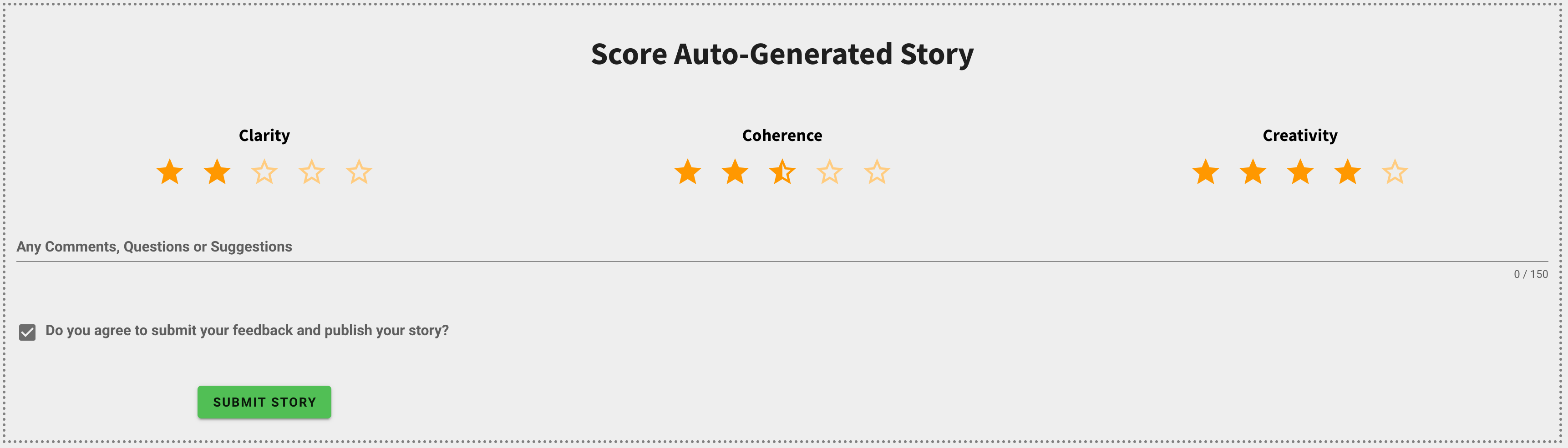}\caption{Auto-Generated Story Feedback Form}\end{subfigure}
    \begin{subfigure}[B]{\textwidth}\includegraphics[width=\linewidth]{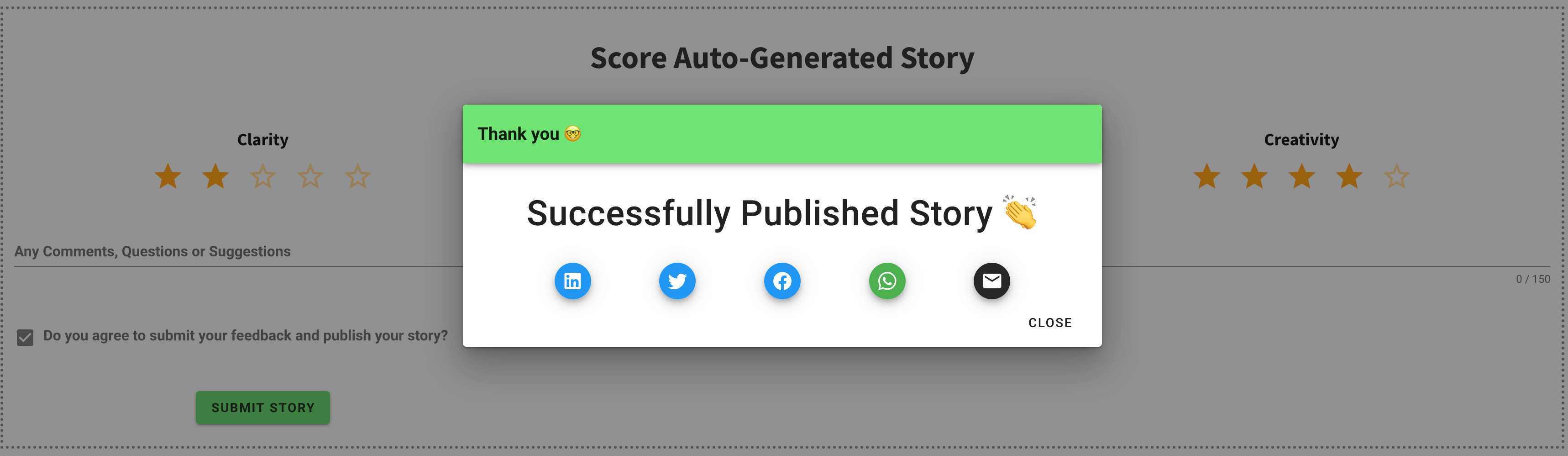}\caption{Published Story is Shareable}\end{subfigure}%
    \caption{Users can publish their created stories, give feedback and share stories with others.}
    \label{fig:feedback}%
\end{Figure}

\section{Feedback from Story Generation Experts }\label{appf-evaluations}
        \begin{itemize}
        \item  \textit{Controllable Neural Story Plot Generation via Reinforcement Learning} \citep{rl-story-generation} co-author, Professor Mark Riedl: The main innovation is the refined, web-based interactive demo that works with writers to create stories. 
        \item \textit{STORIUM: A Dataset and Evaluation Platform for Machine-in-the-Loop Story Generation} \citep{storium} first author, Nader Akoury: The main difference is the  reachability of FairyTailor to any writer, outside of the STORIUM platform, and the ability to write narratives and stories' of any structure. In addition, STORIUM autocompletes do not suggest images. 
        \item \textit{Creative Writing with a Machine in the Loop: Case Studies on Slogans and Stories} \citep{creative-writing-cases} first author, Elizabeth Clark: The significant distinctions are the writing flexibility and the intuitive options menu that enables adding images and texts.   
        \item \textit{Creative Help: A Story Writing Assistant} \citep{creative-help} first-author, Dr. Melissa Roemmele: The dominant strength is the analysis platform that can be used for evaluation thanks to the editable autocompletes and user-inputted content. The images are also novel in comparison to existing story generation platforms. 
        \end{itemize}
\end{document}